\documentclass[11pt]{article}

\usepackage[preprint]{acl}

\usepackage{times}
\usepackage{latexsym}

\usepackage[T1]{fontenc}

\usepackage[utf8]{inputenc}

\usepackage{microtype}

\usepackage{inconsolata}

\usepackage{graphicx}

\usepackage{graphicx}
\usepackage{amsmath,amssymb}
\usepackage{booktabs}
\usepackage{multirow}
\usepackage{caption}
\usepackage{subcaption}
\usepackage{microtype}
\usepackage{url} 

\title{Can Small Language Models Use What They Retrieve?\\
An Empirical Study of Retrieval Utilization Across Model Scale}

\author{
Sanchit Pandey \\
BITS Pilani, Hyderabad Campus \\
Telangana, India \\
\texttt{f20220740@hyderabad.bits-pilani.ac.in}
}

\begin{document}
\maketitle
\begin{abstract}
Retrieval-augmented generation (RAG) is widely deployed to improve factual
accuracy in language models, yet it remains unclear whether smaller models
(${\leq}$7B parameters) can effectively utilize retrieved information.
To investigate this question, we evaluate five model sizes (360M--8B) from three architecture families (SmolLM2, Qwen2.5, Llama~3.1) under four retrieval conditions: no retrieval, BM25, dense (E5-large-v2), and \emph{oracle} retrieval in which the retrieved passages are guaranteed to contain the answer, allowing us to isolate utilization behavior from retrieval accuracy.
We introduce a parametric knowledge split that separates questions a model can already answer from those that require external knowledge, allowing us to isolate utilization failure from retrieval quality failure.
We find three key results.
First, even with oracle retrieval, models ${\leq}$7B fail to extract the
correct answer 85--100\% of the time on questions they cannot answer alone,
indicating a fundamental utilization bottleneck.
Second, any retrieval---including oracle---destroys 42--100\% of previously
correct parametric answers, suggesting a \emph{distraction effect} driven by the mere
presence of context rather than its quality.
Third, error analysis of 2,588 oracle failures shows that the dominant failure
mode is irrelevant generation (61--100\%), where the model ignores the provided context completely.
The pattern is robust across three prompt templates and three retrieval
methods.
Taken together these results indicate that for sub-7B models, the RAG bottleneck is
context utilization, not retrieval quality, and that deploying RAG at this
scale produces a net negative trade-off under standard evaluation
conditions.
\end{abstract}

\section{Introduction}
\label{sec:intro}

Retrieval-augmented generation (RAG) has become a standard technique for
improving factual accuracy in language models
\citep{lewis-etal-2020-retrieval,guu-etal-2020-realm}.
By retrieving relevant passages from an external corpus and conditioning
generation on them, RAG systems can reduce hallucination and provide
up-to-date information.
The appeal of combining small, efficient models with powerful retrieval is
particularly strong for deployment on edge devices, in cost-constrained
settings, and for privacy-sensitive applications.

However, existing RAG evaluations focus primarily on large models ($>$10B parameters), leaving an important question largely unexplored: \emph{can smaller language models actually make effective use of retrieved information?}

Prior work has shown that irrelevant context degrades LLM performance
\citep{shi-etal-2023-large,yoran-etal-2024-making}, but this conflates two
distinct failure modes: (1)~the retriever returning unhelpful passages, and
(2)~the model failing to utilize helpful passages.
\citet{mallen-etal-2023-trust} investigate when retrieval helps based on
entity popularity, but do not isolate the model's ability to extract from
context.
To separate retrieval quality from utilization, we introduce an \emph{oracle} condition in which the provided passages are guaranteed to contain the answer---an experimental setting that has not been systematically examined across model scales.

Our goal in this research is diagnostic.
Instead of proposing a new RAG method, we aim to measure the utilization bottleneck quantitatively, providing an empirically grounded target for future work on fine-tuning, architectural design, or selective retrieval.

To study this question, we evaluate five model sizes (360M to 8B parameters) from three architecture families under four retrieval conditions (none, BM25, dense, oracle) on 1,000 questions drawn from Natural Questions and HotpotQA.
The experimental design includes three methodological components:
(1)~oracle retrieval drawn from the same corpus used by evaluated retrievers,
eliminating corpus distribution mismatch and cleanly isolating utilization
from retrieval quality;
(2)~a parametric knowledge split that classifies each (model, question) pair
as \textsc{Known} or \textsc{Unknown}, enabling separate analysis of
retrieval benefit and retrieval harm; and
(3)~systematic robustness checks across retrieval methods (BM25, dense,
hybrid), prompt templates (forced, permissive, minimal), and question types
(factoid, multi-hop).

Using oracle retrieval together with parametric knowledge splits, we conduct a scaling study that isolates retrieval utilization from retrieval quality.
Our key findings are:
(a)~even with perfect retrieval, the 7B model extracts the correct answer
only 14.6\% of the time for \textsc{Unknown} questions;
(b)~retrieval destroys 42--100\% of correct answers on \textsc{Known}
questions;
(c)~error analysis reveals the dominant failure is irrelevant generation; and
(d)~the pattern holds across all retrieval methods and prompt templates
tested.

\section{Related Work}
\label{sec:related}

\paragraph{Retrieval-augmented generation.}
\citet{lewis-etal-2020-retrieval} and \citet{guu-etal-2020-realm} introduce
RAG for knowledge-intensive tasks, typically evaluating on large models.
\citet{izacard-grave-2021-leveraging} show that Fusion-in-Decoder benefits
from more passages, but evaluate models ${\geq}$3B with task-specific
fine-tuning.
In contrast, we study the under-explored sub-7B regime using instruction-tuned (rather than fine-tuned) models.
\citet{asai-etal-2024-selfrag} propose Self-RAG with self-reflection tokens
to decide when to retrieve.
\citet{jiang-etal-2023-flare} introduce FLARE, which generates forward-looking
retrieval queries during generation.
Our results bring into question a key assumption of such systems: that the model can
effectively use passages once retrieved.
At smaller model scales, the primary bottleneck lies in the ``use'' step itself; selective retrieval may reduce distraction but does not resolve the underlying utilization failure.

\paragraph{Context distraction.}
\citet{shi-etal-2023-large} demonstrate that irrelevant context degrades LLM
performance, but evaluate only large models ($>$10B) and do not include an
oracle condition.
\citet{yoran-etal-2024-making} study making RAG robust to irrelevant context
but do not isolate parametric vs.\ retrieved knowledge.
The oracle setting in our experiments isolates utilization from distraction, revealing that even \emph{relevant} context can degrade performance at smaller scales.
\citet{liu-etal-2024-lost} show that language models struggle to utilize
information in the middle of long contexts, motivating our attention dilution
hypothesis.
\citet{xiong-etal-2023-effective} investigate long-context capabilities in
smaller models, finding significant degradation below 7B parameters.

\paragraph{Parametric vs.\ non-parametric knowledge.}
\citet{mallen-etal-2023-trust} investigate when retrieval helps based on
entity popularity, operating at the entity level.
Our parametric split operates at the individual (model, question) level,
providing finer-grained analysis.
\citet{longpre-etal-2021-entity} study parametric--retrieved knowledge
conflicts in large models; we extend this to the small-model regime.

\paragraph{RAG fine-tuning for small models.}
\citet{zhang-etal-2024-raft} propose RAFT (Retrieval-Augmented Fine-Tuning),
showing that training models specifically for RAG significantly improves
utilization.
\citet{shi-etal-2024-replug} study retrieval-augmented language model
pre-training.
These results reinforce our finding that standard instruction-tuning is
insufficient for grounding at small scales.

\paragraph{Faithfulness and grounding evaluation.}
\citet{min-etal-2023-factscore} introduce FActScore for fine-grained
factuality evaluation.
\citet{es-etal-2023-ragas} propose RAGAS, a framework specifically designed
to evaluate RAG pipelines along multiple dimensions including faithfulness.
Our EM-based evaluation complements these frameworks; we discuss the relationship in Section~\ref{sec:setup}.

\paragraph{Oracle probing.}
\citet{petroni-etal-2021-kilt} use gold passages for knowledge probing in
KILT.
We build on this idea with a systematic scaling study incorporating parametric splits, multiple retrieval baselines, and an error taxonomy designed to reveal the mechanism of failure.

\paragraph{Small model capabilities.}
Recent small-model releases such as Phi-3 \citep{abdin-etal-2024-phi3} and Gemma \citep{team-etal-2024-gemma} demonstrate strong per-parameter performance. This makes the utilization bottleneck we identify particularly notable: even capable small models struggle to ground their answers in retrieved context.

\section{Experimental Setup}
\label{sec:setup}

\subsection{Models}

We evaluate five instruction-tuned models spanning approximately 22$\times$
in parameter count from three architecture families.
The four local models use identical 4-bit NF4 quantization via BitsAndBytes
with double quantization to ensure fair comparison.
Llama-3.1-8B-Instruct is evaluated via Groq API inference (FP16), allowing a cross-precision validation.
The qualitative pattern—where no retrieval outperforms all retrieval methods—matches that observed in the 4-bit local models, indicating that the results are not artifacts of quantization.
All five models are decoder-only, instruction-tuned transformers.

\begin{table}[t]
\centering
\small
\begin{tabular}{lrcc}
\toprule
\textbf{Model} & \textbf{Params} & \textbf{Quant.} & \textbf{Backend} \\
\midrule
SmolLM2-360M-Inst.   & 360M & NF4 4-bit & Local T4 \\
Qwen2.5-1.5B-Inst.   & 1.5B & NF4 4-bit & Local T4 \\
Qwen2.5-3B-Inst.     & 3.0B & NF4 4-bit & Local T4 \\
Qwen2.5-7B-Inst.     & 7.0B & NF4 4-bit & Local T4 \\
Llama-3.1-8B-Inst.$^\dagger$ & 8.0B & FP16 & Groq API \\
\bottomrule
\end{tabular}
\caption{Models evaluated. $^\dagger$Llama-3.1-8B uses FP16 via API;
absolute values not directly comparable, but the
qualitative pattern is consistent (see Section~\ref{sec:cross_arch}).}
\label{tab:models}
\end{table}

\subsection{Datasets}
\label{sec:datasets}

The evaluation set consists of 500 questions each from Natural Questions Open
\citep[NQ;][]{kwiatkowski-etal-2019-natural} and HotpotQA
\citep{yang-etal-2018-hotpotqa}, totalling 1,000 evaluation questions.
NQ provides single-hop factoid questions; HotpotQA provides multi-hop
reasoning questions.
The observed effect sizes (40--65pp distraction, $p < 0.001$ after
Bonferroni correction) provide ample statistical power for all pairwise
comparisons made, consistent with prior controlled RAG studies of similar
scale \citep{mallen-etal-2023-trust}.

We also constructed a PopQA long-tail evaluation set (500 questions), but
our 500K-passage corpus achieved 0\% hit rate across all retrieval methods,
reflecting the challenge of covering long-tail entities with a sub-million
passage corpus.

\subsection{Retrieval Systems}

All retrieval systems---including the oracle---retrieve from the same 500K-passage
Wikipedia corpus (${\approx}$2\% of English Wikipedia, sampled via
deterministic hashing of article titles, chunked at 100 words per passage).
This eliminates corpus distribution mismatch between conditions.

\paragraph{BM25.} We use the bm25s library with English stopword removal,
achieving hit@5 of 10.5\% overall.

\paragraph{Dense (E5-large-v2).} We encode all 500K passages with
\texttt{intfloat/e5-large-v2} (1024-dim embeddings) and index with FAISS
IndexFlatIP.
Dense retrieval achieves hit@5 of 16.3\% overall.

\paragraph{Hybrid (RRF).} Reciprocal Rank Fusion \citep[$k{=}60$;][]{cormack-etal-2009-reciprocal}
combines BM25 and dense rankings, achieving hit@5 of 14.9\%.
Notably, hybrid underperforms pure dense retrieval, suggesting the BM25
signal adds noise at this corpus scale.

\paragraph{Oracle.} For each question, we identify the passage in the same
500K corpus containing a gold answer string, placed at rank 1.
This achieves 100\% hit@1 for the 756 questions where the corpus contained a
matching passage (75.6\% coverage).
We report primary results on these corpus-matched questions only, excluding
244 questions that would require synthetic passages.
A sensitivity analysis excluding the lowest-scoring 15\% of oracle passages
yielded qualitatively identical results (${\leq}$1pp EM change for all models).

\begin{table}[t]
\centering
\small
\begin{tabular}{lrrrr}
\toprule
\textbf{Method} & \textbf{@1} & \textbf{@5} & \textbf{@10} & \textbf{@20} \\
\midrule
BM25         &  4.9 & 10.5 & 13.1 & 16.9 \\
Dense        &  8.7 & 16.3 & 19.5 & 22.9 \\
Hybrid (RRF) &  6.2 & 14.9 & 19.0 & 21.9 \\
Oracle       & 100.0 & 100.0 & 100.0 & 100.0 \\
\bottomrule
\end{tabular}
\caption{Retrieval hit rates (\%) on the 1,000-question evaluation set
(NQ + HotpotQA).}
\label{tab:hitrates}
\end{table}

\subsection{Parametric Knowledge Split}

Each (model, question) pair is classified by first running the model under the \texttt{none} condition.
Correctly answered questions (EM\,=\,1) are labeled \textsc{Known} and
others \textsc{Unknown}.
This yields model-specific splits: SmolLM2-360M knows 1.3\%,
Qwen2.5-1.5B knows 9.0\%, Qwen2.5-3B knows 13.6\%, and Qwen2.5-7B knows
18.5\%.
Note: the \textsc{Known} set for SmolLM2-360M contains only 11 questions;
its distraction effect estimates should be interpreted with extreme caution.

\subsection{Evaluation Protocol}

Exact Match (EM) with standard normalization (lowercasing, article
removal, punctuation stripping) serves as our primary metric, with F1 as a secondary
check.
All results include 95\% bootstrap confidence intervals (2,000 resamples).
Pairwise comparisons use McNemar's test with Bonferroni correction across 24
comparisons ($\alpha = 0.002$).
EM penalises verbose or paraphrased correct answers; our error analysis
accounts for this (verbose correct + partial match = 7--8\% of failures),
and relaxed EM improves scores modestly (${\sim}$6pp for 7B) but does not
change qualitative conclusions.
These evaluation choices are complementary to frameworks such as RAGAS
\citep{es-etal-2023-ragas} and FActScore \citep{min-etal-2023-factscore},
which we leave as future work.

\subsection{Reproducibility}

The code-base is available at \url{https://anonymous.4open.science/r/rag-utilization-study-C67F}.
Experiments were conducted on free-tier Kaggle T4 GPU instances.
The evaluation dataset and oracle passage mappings will be released to
facilitate replication.

\section{Results}
\label{sec:results}

\subsection{Pilot Study: Retrieval Hurts Across All Methods}

Before the oracle experiment, we conducted a pilot study ($n{=}200$;
100 NQ + 100 HotpotQA) comparing four models across three retrieval methods
and a no-retrieval baseline.

\begin{table}[t]
\centering
\small
\begin{tabular}{lrrrr}
\toprule
\textbf{Model} & \textbf{None} & \textbf{BM25} & \textbf{Dense} & \textbf{Hybrid} \\
\midrule
SmolLM2-360M & 1.5  & 0.0 & 0.0  & 0.0 \\
Qwen2.5-1.5B & 14.5 & 5.5 & 3.5  & 6.0 \\
Qwen2.5-3B   & 19.5 & 5.5 & 12.0 & 6.5 \\
Qwen2.5-7B   & 16.5 &  -- & 14.0 &  -- \\
\bottomrule
\end{tabular}
\caption{Pilot study EM (\%) across retrieval methods ($n{=}200$). No
retrieval outperforms all retrieval methods for every model.}
\label{tab:pilot}
\end{table}

Across all models, the no-retrieval baseline outperforms every retrieval method tested.
The $\Delta$EM (dense $-$ none) was negative at every scale:
$-1.5$pp (360M), $-11.0$pp (1.5B), $-7.5$pp (3B), $-2.5$pp (7B).
These observations raise a central question: is the problem that retrieval itself is noisy, or that small models struggle to use even high-quality passages?
The oracle experiment is designed to address this distinction directly.

\subsection{Oracle Retrieval Fails for Small Models}

\begin{figure*}[t]
  \includegraphics[width=\linewidth]{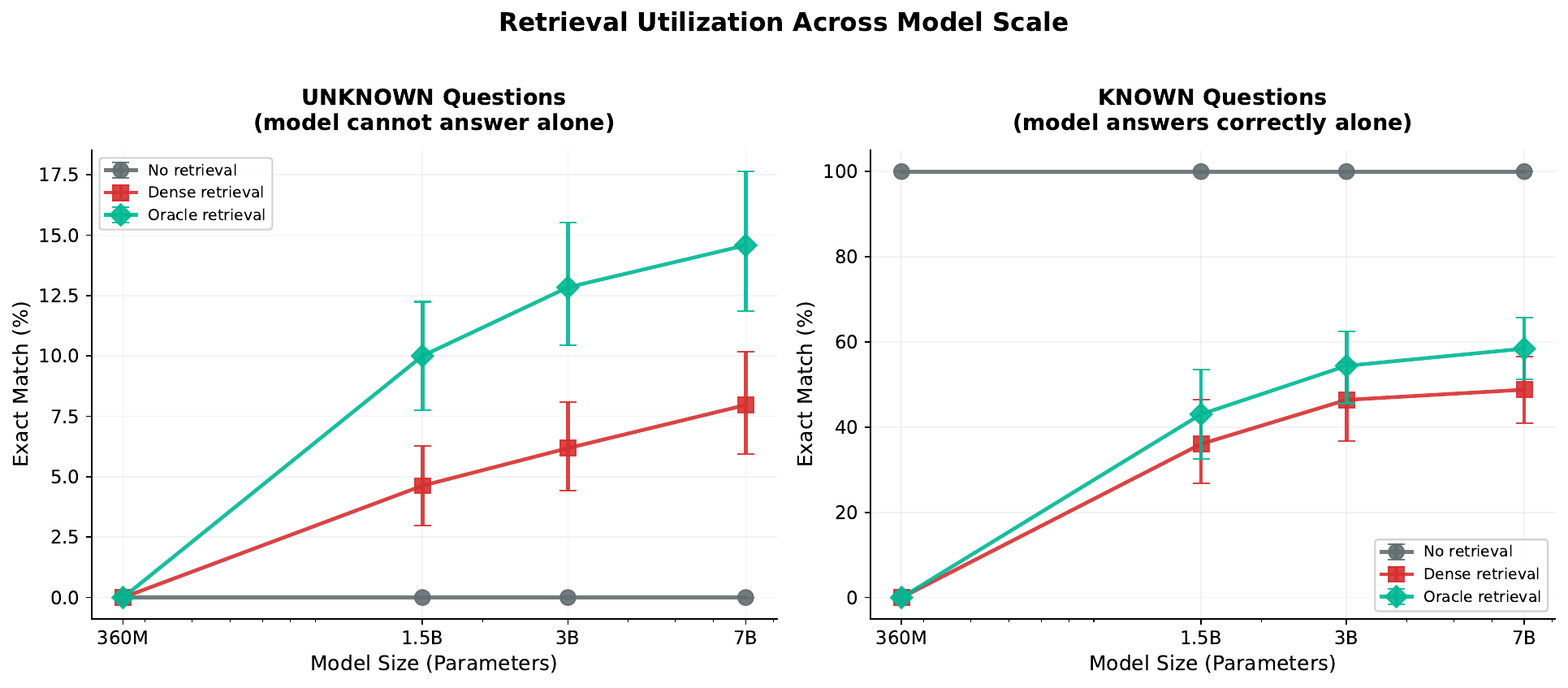}
  \caption{Retrieval utilization across model scale. \textbf{Left:}
    For \textsc{Unknown} questions, oracle retrieval achieves at most 14.6\% EM
    at 7B, meaning 85\%+ of retrieval effort is wasted.
    \textbf{Right:} For \textsc{Known} questions, all retrieval methods destroy
    42--64\% of previously correct answers. Error bars are 95\% bootstrap CIs.}
  \label{fig:scaling}
\end{figure*}

Table~\ref{tab:main} and Figure~\ref{fig:scaling} present our primary results
on corpus-matched questions ($n{=}756$).

\begin{table*}[t]
\centering
\small
\begin{tabular}{llrrrr}
\toprule
\textbf{Split} & \textbf{Retrieval}
  & \textbf{360M} & \textbf{1.5B} & \textbf{3B} & \textbf{7B} \\
\midrule
\multirow{3}{*}{\textsc{Unknown}}
  & None              & 0.0               & 0.0               & 0.0               & 0.0 \\
  & Noisy (dense)     & 0.0               & 4.6 [3.0--6.3]    & 6.2 [4.4--8.1]    & 8.0 [5.9--10.2] \\
  & Oracle            & 0.0               & 10.0 [7.8--12.2]  & 12.8 [10.5--15.5] & 14.6 [11.9--17.6] \\
\midrule
\multirow{3}{*}{\textsc{Known}}
  & None              & 100.0             & 100.0             & 100.0             & 100.0 \\
  & Noisy (dense)     & 0.0               & 36.0 [26.7--46.5] & 46.4 [36.8--55.2] & 48.8 [41.0--56.6] \\
  & Oracle            & 0.0               & 43.0 [32.6--53.5] & 54.4 [45.6--62.4] & 58.4 [51.2--65.7] \\
\bottomrule
\end{tabular}
\caption{Exact Match (\%) by parametric split, retrieval condition, and model
size [95\% CI]. Corpus-matched oracle only ($n{=}756$). All non-zero
differences from None are significant at $p < 0.001$ (McNemar,
Bonferroni-corrected), except SmolLM2-360M. The 360M \textsc{Known} cell
is based on $n{=}11$ only and should be interpreted with extreme caution.}
\label{tab:main}
\end{table*}

For \textsc{Unknown} questions, the oracle condition guarantees that the answer appears in the passage, yet the 7B model achieves only 14.6\% EM [11.9--17.6].
The 1.5B model manages 10.0\% [7.8--12.2], and the 360M scores 0.0\%
across all conditions.
Even with a perfect retrieval system, 85--100\% of retrieval effort is
wasted for these models.
Oracle retrieval roughly doubles the benefit compared to noisy retrieval
(14.6\% vs.\ 8.0\% for 7B), confirming that retrieval quality does
matter---but even perfect retrieval yields very low absolute performance,
demonstrating that the binding constraint is utilization, not retrieval
quality.

\subsection{Retrieval Destroys Parametric Knowledge}

The \textsc{Known} rows of Table~\ref{tab:main} reveal a severe distraction
effect.
Models achieving 100\% EM without retrieval suffer catastrophic drops when
passages are added: the 7B model loses 51.2pp with noisy retrieval and
41.6pp with oracle retrieval; the 1.5B model loses 64.0pp and 57.0pp
respectively.

All distraction effects are significant at $p < 0.001$.
Notably, the difference between oracle and noisy distraction is
\emph{not significant} for models below 7B (1.5B: $p{=}0.26$; 3B:
$p{=}0.07$), suggesting that the \emph{presence} of any context---not its
quality---drives the distraction effect.
Only at 7B does oracle retrieval preserve significantly more knowledge than
noisy retrieval ($p{=}0.003$).

\begin{figure}[t]
  \includegraphics[width=\columnwidth]{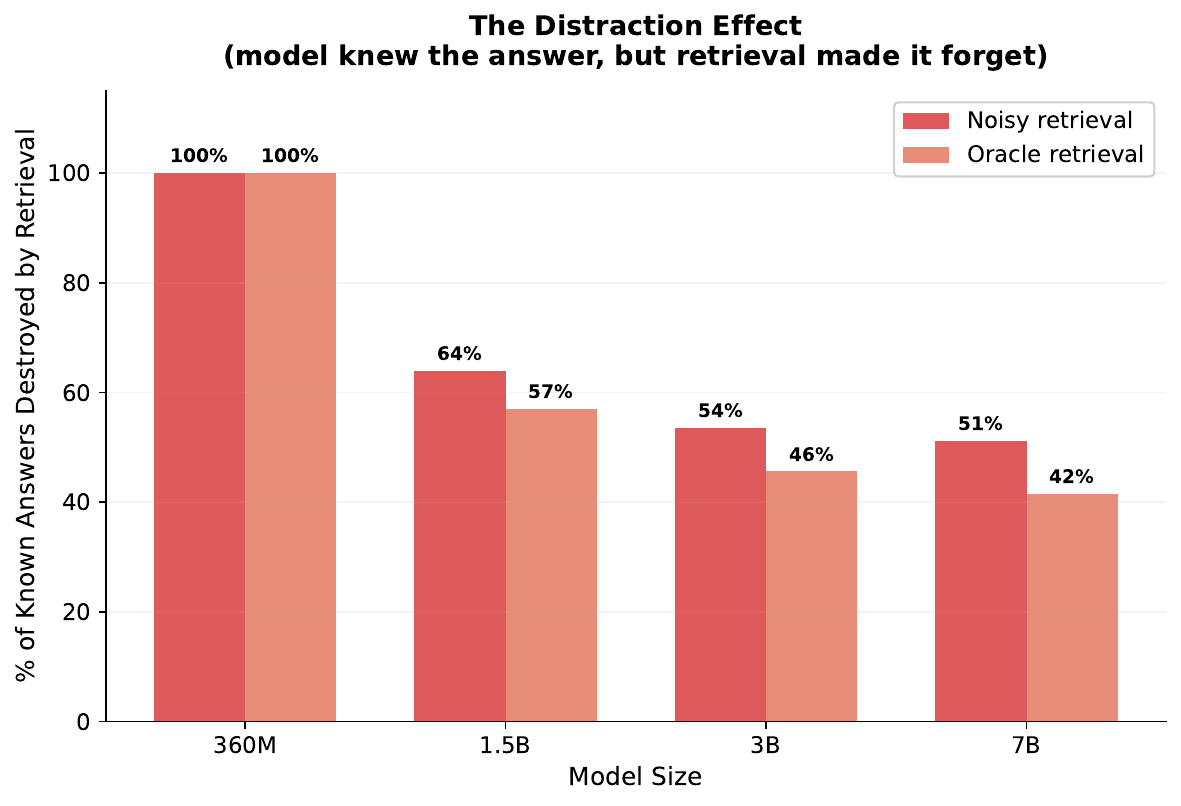}
  \caption{The distraction effect: percentage of \textsc{Known} answers
    destroyed by adding retrieval context. For models ${\leq}$3B, oracle and
    noisy retrieval cause statistically indistinguishable harm---context
    presence, not quality, drives distraction.}
  \label{fig:distraction}
\end{figure}

\subsection{The Net Trade-off Is Negative}

The expected net EM change from adding retrieval is:
\begin{equation}
\Delta\mathrm{EM}_\mathrm{net} =
  p_\mathrm{unk} \cdot \Delta\mathrm{EM}_\mathrm{unk}
+ p_\mathrm{kn} \cdot \Delta\mathrm{EM}_\mathrm{kn}
\end{equation}

\begin{table}[t]
\centering
\small
\begin{tabular}{lrrrr}
\toprule
\textbf{Model} & $p_\mathrm{kn}$ & $\Delta$\textsc{Unk} & $\Delta$\textsc{Kn} & \textbf{Net} \\
\midrule
360M$^*$ & 0.013 & +0.0 & $-$100.0 & $-$1.3 \\
1.5B     & 0.090 & +4.6 & $-$64.0  & $-$1.6 \\
3B       & 0.136 & +6.2 & $-$53.6  & $-$1.9 \\
7B       & 0.185 & +8.0 & $-$51.2  & $-$2.9 \\
\bottomrule
\end{tabular}
\caption{Net EM trade-off for noisy (dense) retrieval. The net effect is
negative for every model tested. $^*$360M: $p_\mathrm{kn}$ based on
$n{=}11$; interpret with caution.}
\label{tab:net}
\end{table}

For the 7B model: $(0.815 \times 8.0) - (0.185 \times 51.2) =
+6.5 - 9.5 = -3.0$ percentage points.
Even with oracle retrieval, the 7B model nets only $+4.2$pp---a modest
gain requiring perfect retrieval to achieve.

\subsection{Per-Dataset Analysis}

The pattern is consistent across both datasets but with notable differences.
Oracle utilization on \textsc{Unknown} questions is slightly higher on
HotpotQA for larger models (16.9\% vs.\ 12.5\% for 7B), possibly because
multi-hop questions have more distinctive answer strings that are easier to
locate within the passage.
The distraction effect is stronger for HotpotQA at 1.5B (70.4\% destruction
vs.\ 53.1\% for NQ), suggesting multi-hop context is more confusing for
smaller models.
Despite these differences, the qualitative pattern---negative net
trade-off, low utilization, severe distraction---holds for both datasets.

\subsection{Prompt Robustness}

One possible concern is that the results might depend on the specific prompt template used.
We tested three prompt variants on Qwen2.5-3B with dense retrieval on the
200-question pilot set:

\textbf{v1 (forced):} ``Answer based on the provided context.'' (default)\\
\textbf{v2 (permissive):} ``Use context only if it helps; otherwise rely on
your own knowledge.''\\
\textbf{v3 (minimal):} Question with ``Reference material'' appended, no
instruction to use it.

\begin{table}[t]
\centering
\small
\begin{tabular}{lrrr}
\toprule
\textbf{Prompt} & \textbf{Overall EM} & \textbf{In ctx.} & \textbf{Not in ctx.} \\
\midrule
v1 (forced)     & 12.0 & 27.1 & 3.8 \\
v2 (permissive) & 12.5 & 28.6 & 3.8 \\
v3 (minimal)    &  9.0 & 18.6 & 3.8 \\
No retrieval    & 19.5 & 37.1 & 10.0 \\
\bottomrule
\end{tabular}
\caption{Prompt ablation on Qwen2.5-3B $\times$ dense ($n{=}200$). No prompt
recovers the no-retrieval baseline (19.5\% EM).}
\label{tab:prompt}
\end{table}

The permissive prompt (v2), which explicitly allows the model to ignore
context, performs nearly identically to the forced prompt (12.5\% vs.\
12.0\%).
Even when told it may rely on its own knowledge, the 3B model still performs
7pp worse than no retrieval.
None of the three approaches recovers the no-retrieval baseline of 19.5\%.

\subsection{Cross-Architecture Validation}
\label{sec:cross_arch}

To test whether this behavior generalizes beyond the SmolLM2 and Qwen2.5 families, we evaluated Llama-3.1-8B-Instruct. (via Groq API, FP16) on 200 NQ questions under all
three retrieval methods.
Despite being a different architecture family and using full-precision
inference, Llama-3.1-8B exhibits the same pattern: every retrieval method
reduces accuracy relative to no retrieval.

\begin{table}[t]
\centering
\small
\begin{tabular}{lrr}
\toprule
\textbf{Retrieval} & \textbf{EM (\%)} & \textbf{F1 (\%)} \\
\midrule
None          & 24.0 & 34.5 \\
BM25          & 16.5 & 24.2 \\
Dense         & 17.5 & 25.3 \\
Hybrid        & 16.5 & 25.2 \\
\bottomrule
\end{tabular}
\caption{Llama-3.1-8B-Instruct (FP16, Groq API) on 200 NQ questions.
Same qualitative pattern: none outperforms all retrieval methods.
Not directly comparable to 4-bit local models in absolute terms.}
\label{tab:llama}
\end{table}

The $\Delta$EM ranges from $-6.5$pp (dense) to $-7.5$pp (BM25/hybrid),
consistent with the Qwen2.5-7B results.
This cross-architecture, cross-precision replication substantially
strengthens our claim that the utilization bottleneck is not family-specific.

\section{Error Analysis}
\label{sec:error}

To find out why models fail on oracle questions, we classify all 2,588
corpus-matched oracle failures across four local models into six categories
using automated heuristics.

\begin{figure}[t]
  \includegraphics[width=\columnwidth]{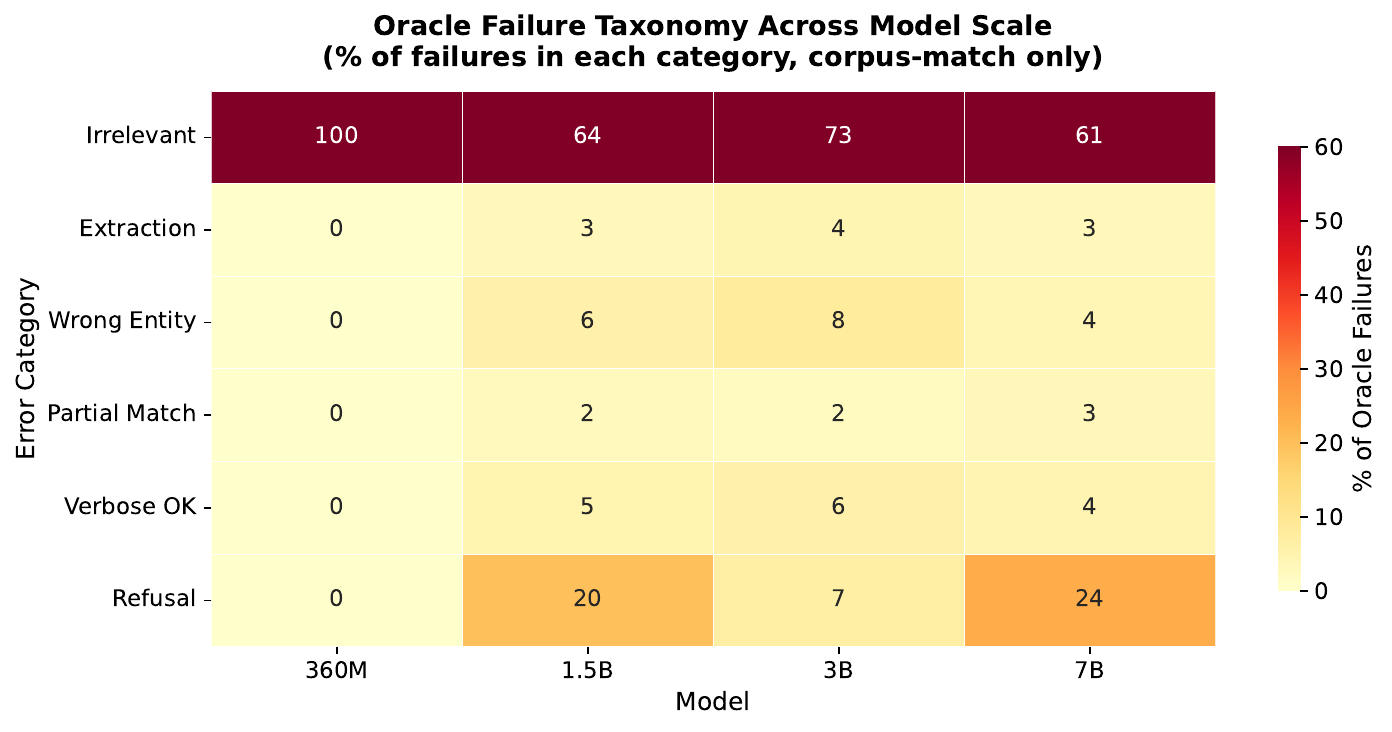}
  \caption{Oracle failure taxonomy (\% of failures per category,
    corpus-matched only, $n{=}2{,}588$). Irrelevant generation is dominant
    at all scales (61--100\%). Refusal is unexpectedly high at 7B (24\%).
    Inter-annotator agreement: 89\% (96 manually reviewed samples).}
  \label{fig:errors}
\end{figure}

\paragraph{Irrelevant generation} is the most frequent failure mode, accounting for 100\% of 360M failures, 64\% at 1.5B, 73\% at 3B, and 61\% at 7B.
The model generates text with no meaningful overlap with the provided passage,
suggesting it ignores the context entirely.

\paragraph{Refusal} appears the second largest category for
instruction-tuned models (20\% at 1.5B, 24\% at 7B), where models respond
with hedging language despite the answer being present.
The 3B model shows substantially lower refusal (7\%) than both 1.5B and 7B,
likely reflecting differences in instruction tuning recipes across model
sizes within the Qwen2.5 family rather than a genuine scaling property.

\paragraph{Format-related failures} (partial match + verbose correct) account
for 7--8\% of errors.
Adjusting EM to count these as correct yields modest improvements (e.g., 7B
oracle on \textsc{Unknown}: 14.6\% $\to$ ${\sim}$21\%), but the fundamental
utilization gap remains large.

A qualitative shift occurs between 360M and 1.5B: 360M produces completely
incoherent outputs, while models ${\geq}$1.5B at least attempt to engage with
the passage.
This suggests a minimum capability threshold (${\sim}$1B parameters) below
which RAG is completely nonfunctional.

\section{Discussion}
\label{sec:discussion}

\subsection{The Utilization Bottleneck}

Overall the results find that the primary bottleneck for small-model RAG is not
retrieval quality, but context utilization.
The oracle condition eliminates retrieval noise, yet performance
remains dramatically low.
This re-frames the problem: efforts to improve retrieval for small-model
deployment (better embeddings, hybrid search, re-ranking) address at most
half the gap between noisy and oracle performance---the larger half requires
improving the model's ability to condition on retrieved text.

\subsection{Why Does Context Hurt Known Answers?}

We consider three possible explanations for the distraction effect, each requiring
further investigation:
(1)~\emph{attention dilution}, where longer contexts may reduce the weight
on the question \citep{liu-etal-2024-lost};
(2)~\emph{instruction conflict}, where the prompt to ``answer based on
context'' overrides parametric knowledge; and
(3)~\emph{position bias}, where small models preferentially attend to early
context tokens.
The prompt ablation (Table~\ref{tab:prompt}) provides partial evidence for
hypothesis~(2): even the permissive prompt fails to recover baseline
performance.
We observe that oracle and noisy retrieval cause statistically
similar distraction for models below 7B further supporting that the presence of context, not its quality, drives the harm.

\subsection{Implications for Practitioners}

Our net trade-off analysis (Table~\ref{tab:net}) has direct practical
implications.
For sub-7B models on general-domain QA, RAG frequently reduces rather than
improves overall accuracy under standard prompting conditions.
Practitioners should consider:
(1)~using retrieval only when the model is likely to lack the answer
(adaptive retrieval), conditional on reliable confidence estimation;
(2)~investing in larger models rather than better retrieval infrastructure;
or (3)~training small models specifically for context utilization via
RAG-aware fine-tuning \citep[e.g., RAFT-style;][]{zhang-etal-2024-raft}.

\subsection{What Scale Is Needed?}

Utilization improves roughly log-linearly with scale in our experiments, but remains far below practical levels even at 7B.
Extrapolating the oracle utilization curve (0\% at 360M $\to$ 10\% at
1.5B $\to$ 13\% at 3B $\to$ 15\% at 7B), robust context utilization
($>$50\% oracle EM) may require models larger than
7B---consistent with \citet{shi-etal-2023-large}'s finding that distraction
resistance emerges primarily in models $>$10B parameters.

\paragraph{Key takeaway.} Improving retrieval systems alone is unlikely to substantially improve small-model RAG performance unless models themselves become better at conditioning on retrieved context.

\section{Conclusion}
\label{sec:conclusion}

We present a controlled study demonstrating that small language models
(${\leq}$7B parameters) struggle to utilize retrieved information for
extractive question answering.
Even under oracle retrieval---where the passage is guaranteed to contain the
answer---models still fail to extract the correct answer 85--100\% of the time,
with the dominant failure being complete disregard of the provided context.
At the same time, adding retrieval context overturns 42--64\% of answers the
model previously answered correctly, an effect driven by the presence of context rather
than its quality.
This pattern holds across three architecture families, three retrieval methods, and three prompt templates.

Deploying RAG with sub-7B models for extractive QA frequently reduces rather
than improves accuracy under standard prompting conditions.
The bottleneck for small-model RAG is context utilization---a capability that
scales slowly with model size and appears to require substantially more than
7B parameters to emerge robustly.
Several directions may help address this limitation, including RAG-aware fine-tuning, selective retrieval conditioned on model confidence, and architectural modifications aimed at improving context grounding at smaller scales.

\section*{Limitations}

Our conclusions should be interpreted within the scope of extractive
question-answering settings, where grounding success can be measured
objectively via exact answer matching.

\paragraph{Task scope.}
Our evaluation is limited to extractive question answering.
The results may not extend directly to generative tasks such as summarization with retrieved documents, reasoning tasks where retrieved information serves as
background rather than containing a direct answer, or tasks where partial
utilization is sufficient.

\paragraph{Model diversity.}
We evaluate five models from three families (SmolLM2, Qwen2.5, Llama~3.1).
While the pattern across the tested model families suggest generalizability,
results may differ for other architectures (e.g., Phi-3, Gemma).
All five are decoder-only, instruction-tuned transformers; encoder-decoder
models may exhibit different utilization characteristics.

\paragraph{Quantization.}
All local models use 4-bit NF4 quantization.
However, the FP16 Llama-3.1-8B results (Table~\ref{tab:llama}) demonstrate
qualitatively identical behaviour to the 4-bit local models across all
retrieval conditions, providing cross-precision validation of our main
finding without requiring additional local experiments.
Furthermore, the 360M model's near-zero performance without retrieval (1.3\%
EM) suggests that quantization alone cannot explain utilization failure.

\paragraph{Corpus coverage and oracle quality.}
Our corpus covers ${\approx}$2\% of English Wikipedia (500K passages),
limiting oracle coverage to 75.6\%.
Some oracle passages may contain incidental answer mentions.
Expanding coverage would improve oracle quality and enable PopQA evaluation.

\paragraph{Prompt sensitivity.}
Our three-prompt ablation demonstrates robustness of the qualitative pattern,
but is limited to the 200-question pilot set and one model size (3B, with
spot checks on 7B).

\paragraph{Domain and language scope.}
We evaluate English-language, Wikipedia-domain QA only.
Performance may differ on specialized domains or other languages, though the utilization bottleneck should persist or worsen in lower-resource
settings.

\bibliography{references}

\appendix

\section{Retrieval Hit Rates by Dataset}
\label{app:hitrates}

\begin{table}[h]
\centering
\small
\begin{tabular}{llrrrr}
\toprule
\textbf{Method} & \textbf{Dataset} & \textbf{@1} & \textbf{@5} & \textbf{@10} & \textbf{@20} \\
\midrule
BM25   & NQ       &  7.6 & 15.8 & 20.0 & 26.8 \\
BM25   & HotpotQA &  7.2 & 15.6 & 19.4 & 23.8 \\
BM25   & PopQA    &  0.0 &  0.0 &  0.0 &  0.0 \\
Dense  & NQ       & 15.8 & 28.2 & 32.6 & 38.2 \\
Dense  & HotpotQA & 10.2 & 20.8 & 26.0 & 30.6 \\
Dense  & PopQA    &  0.0 &  0.0 &  0.0 &  0.0 \\
Hybrid & NQ       & 11.2 & 24.8 & 31.2 & 35.2 \\
Hybrid & HotpotQA &  7.4 & 20.0 & 25.8 & 30.6 \\
Hybrid & PopQA    &  0.0 &  0.0 &  0.0 &  0.0 \\
\bottomrule
\end{tabular}
\caption{Retrieval hit rates (\%) by dataset.}
\end{table}

\section{Latency Analysis}
\label{app:latency}

\begin{table}[h]
\centering
\small
\begin{tabular}{lr}
\toprule
\textbf{Component} & \textbf{Latency (ms/query)} \\
\midrule
BM25 retrieval            &   3.5 \\
FAISS search (exact IP)   & 180.2 \\
E5-large-v2 query encoding &  ${\sim}$50 \\
Dense total               & ${\sim}$230 \\
Hybrid (BM25+dense+RRF) & ${\sim}$234 \\
\bottomrule
\end{tabular}
\caption{Latency measured on Kaggle T4 GPU with 500K-passage corpus.
BM25 is ${\approx}$66$\times$ faster than dense retrieval.}
\end{table}

\section{Statistical Significance Tests}
\label{app:stats}

All pairwise comparisons use McNemar's test with Bonferroni correction
(24 comparisons, $\alpha = 0.002$).

\paragraph{Oracle vs.\ None (\textsc{Unknown}).}
Significant for 1.5B, 3B, 7B (all $p < 0.001$). Not significant for 360M
($p = 1.0$; 0 correct in both conditions).

\paragraph{Noisy vs.\ None (\textsc{Known} --- distraction).}
Significant for all models ${\geq}$1.5B (all $p < 0.001$).

\paragraph{Oracle vs.\ Noisy (\textsc{Known} --- distraction quality).}
Not significant for 1.5B ($p = 0.26$) or 3B ($p = 0.07$).
Significant for 7B ($p = 0.003$).
This confirms that for models below 7B, distraction is driven by context
presence, not quality.

\section{Pilot Study Full Results}
\label{app:pilot}

\begin{table}[h]
\centering
\small
\begin{tabular}{lrrr}
\toprule
\textbf{Model} & \textbf{None EM} & \textbf{Dense EM} & $\Delta$\textbf{EM} \\
\midrule
SmolLM2-360M & 1.5  & 0.0  & $-$1.5  \\
Qwen2.5-1.5B & 14.5 & 3.5  & $-$11.0 \\
Qwen2.5-3B   & 19.5 & 12.0 & $-$7.5  \\
Qwen2.5-7B   & 16.5 & 14.0 & $-$2.5  \\
\bottomrule
\end{tabular}
\caption{Pilot study full results ($n{=}200$, $\Delta$EM = dense $-$ none).}
\end{table}

\section{Supplementary Figures}
\label{app:figures}

\begin{figure}[h]
  \includegraphics[width=\columnwidth]{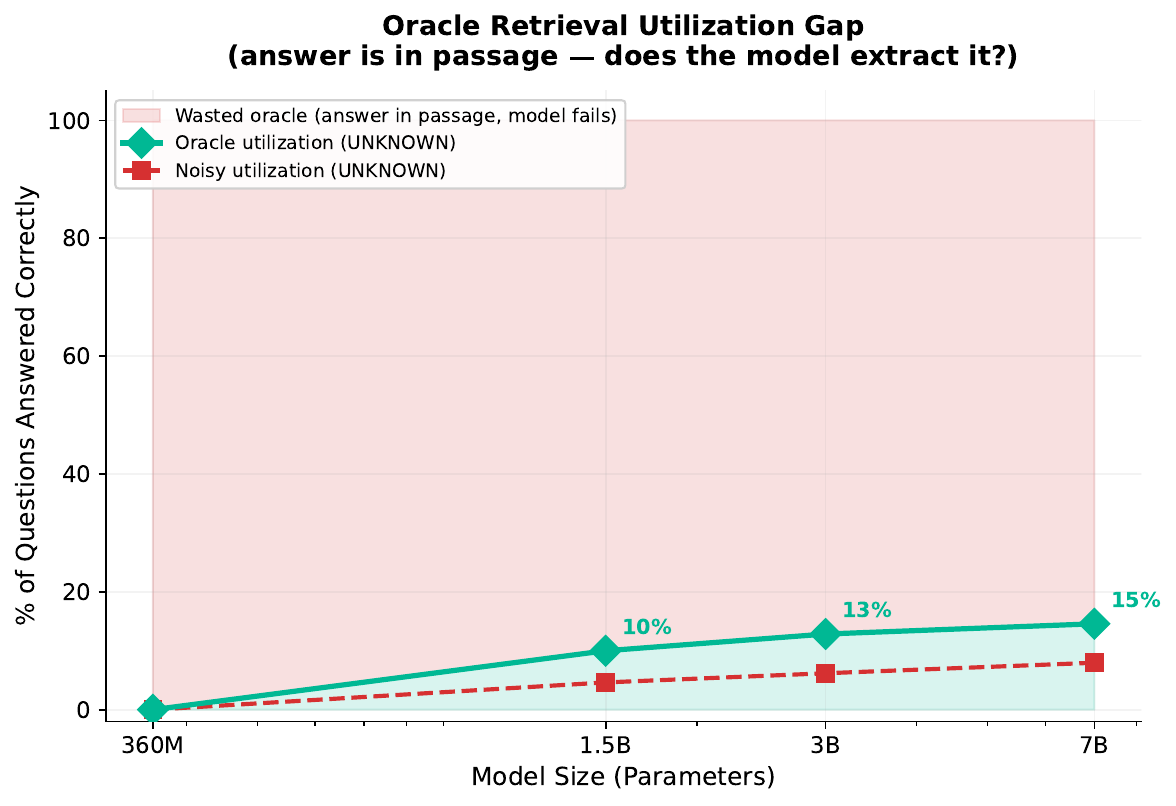}
  \caption{Oracle utilization gap: the pink region shows the fraction of
    retrieval effort that is wasted (answer in passage, model still fails).
    Even at 7B, 85\% of oracle retrievals are unused.}
  \label{fig:gap}
\end{figure}

\begin{figure}[h]
  \includegraphics[width=\columnwidth]{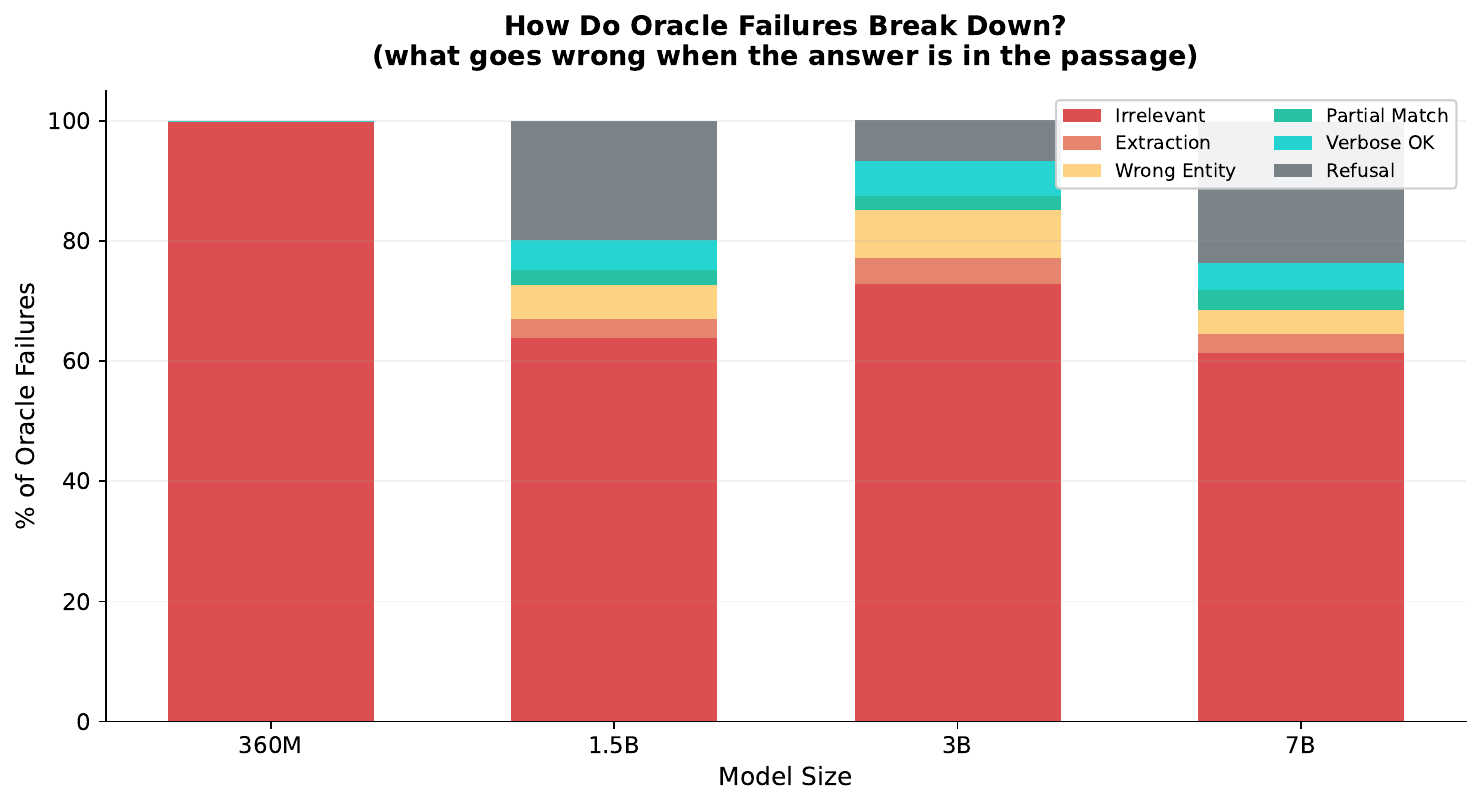}
  \caption{Error category breakdown as stacked proportions.
    Irrelevant generation decreases as a share with scale but remains the
    dominant failure mode. Refusal is non-monotone across Qwen2.5 sizes,
    likely reflecting instruction-tuning recipe differences.}
  \label{fig:stacked}
\end{figure}

\section{Example Error Outputs}
\label{app:errors}

Representative examples from each error category (Qwen2.5-3B, oracle
retrieval).

\paragraph{Irrelevant generation (73\% of 3B failures).}
\textit{Q:} ``What country is the Jutland Peninsula in?''
\textit{Gold:} Denmark.
\textit{Prediction:} ``The Jutland Peninsula is a large body of water that
separates the North Sea from the Baltic Sea.''
The model generates a plausible-sounding but factually wrong response with no
overlap with the passage content.

\paragraph{Refusal (7\% of 3B failures).}
\textit{Q:} ``Who wrote the music for Phantom of the Opera?''
\textit{Gold:} Andrew Lloyd Webber.
\textit{Prediction:} ``Based on the provided context, I cannot determine who
wrote the music for Phantom of the Opera.''
The answer appears in the oracle passage, but the model refuses to extract it.

\paragraph{Wrong entity (8\% of 3B failures).}
\textit{Q:} ``Who directed Schindler's List?''
\textit{Gold:} Steven Spielberg.
\textit{Prediction:} ``Liam Neeson.''
The model extracts a named entity from the passage but selects the wrong one.

\paragraph{Partial match (2\% of 3B failures).}
\textit{Q:} ``What is the capital of Australia?''
\textit{Gold:} Canberra.
\textit{Prediction:} ``The capital of Australia is Canberra, which was
selected as the capital in 1908.''
Correct answer present but embedded in verbose output.

\paragraph{Verbose correct (6\% of 3B failures).}
\textit{Q:} ``When was the Eiffel Tower built?''
\textit{Gold:} 1889.
\textit{Prediction:} ``According to the context, the Eiffel Tower was
constructed between 1887 and 1889 for the 1889 World's Fair in Paris,
France.''
Semantically correct and grounded, but EM counts this as incorrect.

\end{document}